\documentclass{article}
\usepackage{ijcai26}

\usepackage{times}
\usepackage{soul}
\usepackage{url}
\usepackage[hidelinks]{hyperref}
\usepackage[utf8]{inputenc}
\usepackage[small]{caption}
\usepackage{graphicx}
\usepackage{amsmath}
\usepackage{amsthm}
\usepackage{natbib}
\usepackage{subcaption}
\usepackage{booktabs}
\let\cite\citep
\usepackage{booktabs}
\usepackage{algorithm}
\usepackage{algorithmic}
\usepackage[normalem]{ulem}
\usepackage{array} 
\usepackage{booktabs,multirow,tabularx,array}
\PassOptionsToPackage{floatrowsep=qquad}{floatrow}
\usepackage{amsmath}   
\usepackage{amssymb}   
\usepackage{amsfonts}  
\usepackage{multirow}
\usepackage{graphicx}
\usepackage{bm} 

\usepackage{mathtools}
\usepackage[switch]{lineno}

\newcommand{\blfootnote}[1]{%
  \begingroup
  \renewcommand{\thefootnote}{}%
  \footnotetext{#1}%
  \endgroup
}

\title{Multi-Scale Generative Modeling with Heat Dissipation Flow Matching}

\author{
Jun Ma$^{1,3,4,\#}$\and
Hanquan Zhang$^{2,\#}$\and
Yanjun Qin$^{2,*}$\And
Haoyuan Guan$^2$\and
Ke Zhang$^{1,3,4,*}$\\
\affiliations
$^1$Department of Systems Science, Faculty of Arts and Sciences, Beijing Normal University, Zhuhai 519087, China\\
$^2$School of Computer Science and Technology, Xinjiang University, Urumqi 830049, China\\
$^3$International Academic Center of Complex Systems, Beijing Normal University, Zhuhai 519087, China\\
$^4$School of Systems Science, Beijing Normal University, Beijing 100875, China\\
\emails
kezhang@bnu.edu.cn,
qinyanjun@xju.edu.cn
}

\begin{document}

\maketitle
\blfootnote{\textsuperscript{\#} Equal contribution.}
\blfootnote{\textsuperscript{*} Corresponding author.}

\begin{abstract}


Diffusion models are widely used in image generation, with most relying on noise-based corruption and denoising. A distinct branch instead uses blur as the main corruption, preserving better color budgets and multi-scale detail by providing multi-scale priors. However, blur-based models remain in SDE-based frameworks and are not integrated into ODE-based frameworks, such as Flow Matching (FM). Meanwhile, in the blur-based formulation, the classical inverse heat-dissipation (IHD) process faces an ill-posed challenge. Moreover, under the data-manifold assumption, regressing blurred images from high-dimensional noise (or velocity) space is also difficult. We propose Heat Dissipation Flow Matching (HDFM), which introduces a continuous blurred (heat-dissipation) process into FM to inject multi-scale priors. HDFM aligns an interpolated heat-dissipation path to address ill-posedness and adopts $x$-prediction to mitigate high-dimensional regression difficulty. Toy experiments and ablation studies show that HDFM consistently benefits from both blur and $x$-prediction. The performance of HDFM outperforms most baseline methods on all datasets. The core code at the link: https://github.com/majunzd/HDFM.
\end{abstract}

\section{Introduction}


Diffusion models draw growing attention in industry~\cite{bjorck2025gr00t}, and academia ~\cite{watson2023novo}, where most methods operate by iterative noise addition and then denoising~\cite{dhariwal2021diffusion,sohl2015deep,lipman2023flow}. A distinct line of work replaces or augments purely noise-based schemes with more physically inspired image corruption processes, such as Inverse Heat  Dissipation~(IHDM)~\cite{rissanen2023generative} or Blurring Diffusion~\cite{hoogeboom2023blurring}. This branch preserves better color budgets and multi-scale detail by retaining cross-scale structural priors~\cite{gruszczynski2025beyond,bansal2023cold}. Together, these approaches ground core theory and substantially improve generative performance.

Classic noising-based diffusion models (\textit{e.g.}, DDPM~\cite{ho2020denoising}) are built on stochastic differential equation (SDE)~\cite{songscore}, which makes inference slow. Flow Matching (FM)~\cite{lipman2023flow} learns a velocity field and replaces the SDE with an ordinary differential equation (ODE) to speed up sampling. Yet under the data-manifold assumption~\cite{chapelle2006b}, predicting clean data directly in a noise-based velocity space remains challenging. JiT~\cite{li2025jit} addresses this dimensional mismatch by predicting the data itself. However, \textbf{its sampling still suffers from continuous cross-scale issues}: scale-transition errors are hard to supervise, and cross-scale structural priors are weakened.
\begin{figure}[t]
    \centering
    \includegraphics[width=0.5\textwidth]{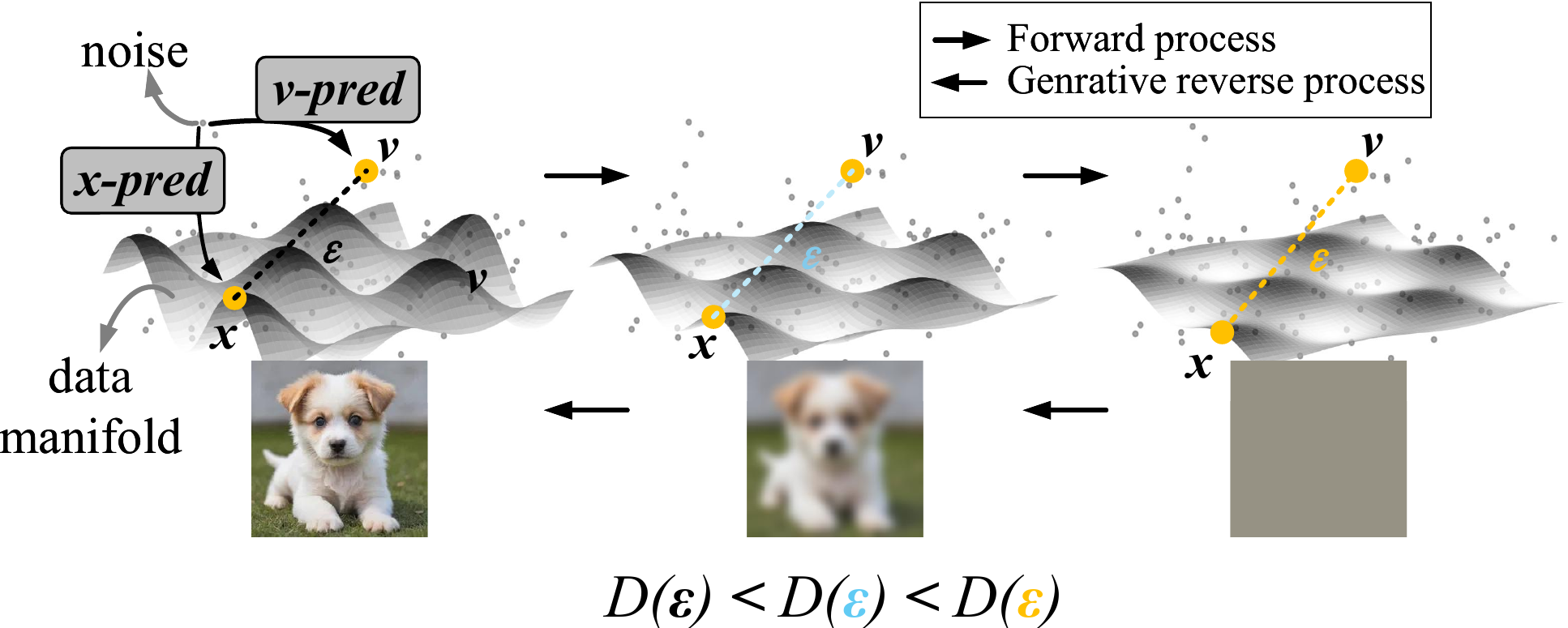}
    \caption{\textbf{Manifold Assumption View of Heat Dissipation Process:} $D(\cdot)$ denotes the gap of dimensionality between the velocity and the data manifold. As forward heat dissipation proceeds, the data-manifold dimension further contracts, while noise and velocity still spread over the full high-dimensional space. Thus, in the heat-dissipation setting, blur image prediction is different from noise or velocity prediction.}
    \label{fig:1}
\end{figure}

For blurring-based diffusion models, IHDM~\cite{rissanen2023generative} handles continuous cross-scale generation only within the SDE framework and does not extend to the FM setting. \textbf{Its reverse process is also ill-posed}~\cite{hoogeboom2023blurring}, which limits sample quality. Moreover, heat dissipation reduces the effective data dimensionality (Fig.~\ref{fig:1}), further widening the gap between the noise-based velocity space and the data space. Under the manifold assumption, \textbf{predicting heat-dissipation blur data from high-dimensional noise is likewise difficult}.


In this paper, we propose \textit{Heat Dissipation Flow Matching (HDFM)}, which addresses these challenges with two key technical components. HDFM forms explicit interpolation coarse-to-fine continuous probabilistic paths through heat-time calibration, which brings multi-scale structural priors to FM while solving the ill-posed inverse heat dissipation problem of IHDM. HDFM specialize $x$-prediction of JiT for the heat-dissipation setting, avoiding the high-dimensional regression challenge caused by $\epsilon$-prediction and $v$-prediction. This improves the stability and quality.

From an engineering perspective, HDFM introduces a lightweight LayerSync regularizer to enhance representational consistency, and our method is compatible with class-conditional generation and classifier-free guidance (CFG)~\cite{ho2021classifier}. We conduct systematic experiments on some image datasets, reporting performance and ablations. We summarize HDFM contributions as follows:

\paragraph{Contributions.}
\begin{enumerate}
  \item HDFM introduces a continuous cross-scale heat-dissipation path into flow matching and mitigates the ill-posed challenge of the original inverse heat process.
  \item HDFM introduces an $x$-pred training paradigm for heat dissipation, addressing the difficulty of high-dimensional regression under the manifold assumption.
  \item Effectiveness of HDFM is validated through comparisons with baseline algorithms and ablation studies.
\end{enumerate}

\section{Preliminaries}

\subsection{Flow Matching}
Flow Matching (FM)~\cite{lipman2023flow} learns a vector field $v(z,t)$ under the Continuous Normalizing Flow (CNF) framework~\cite{chen2018neural}, and the vector field can be written by the ODE
\begin{equation}
\frac{\mathrm{d} z_t}{\mathrm{d} t}=v(z_t,t),\quad z_0\sim \mathcal{N}(\mu,\sigma),\; z_1\approx p_{\text{data}}.
\end{equation}
FM adopts the linear interpolation path~\cite{liu2023flow} between data and noise, $z_t=t x+(1-t)\epsilon$ with $t\in[0,1]$ and $\epsilon\sim\mathcal{N}(0,I)$, and defines a supervision signal $v$ in the velocity space by minimizing
\begin{equation}
\mathcal{L}=\mathbb{E}_{t,x,\epsilon}\Big[\big\|v_\theta(z_t,t)-v\big\|_2^2\Big].
\end{equation}

\begin{figure}[t]
    \includegraphics[width=0.5\textwidth]{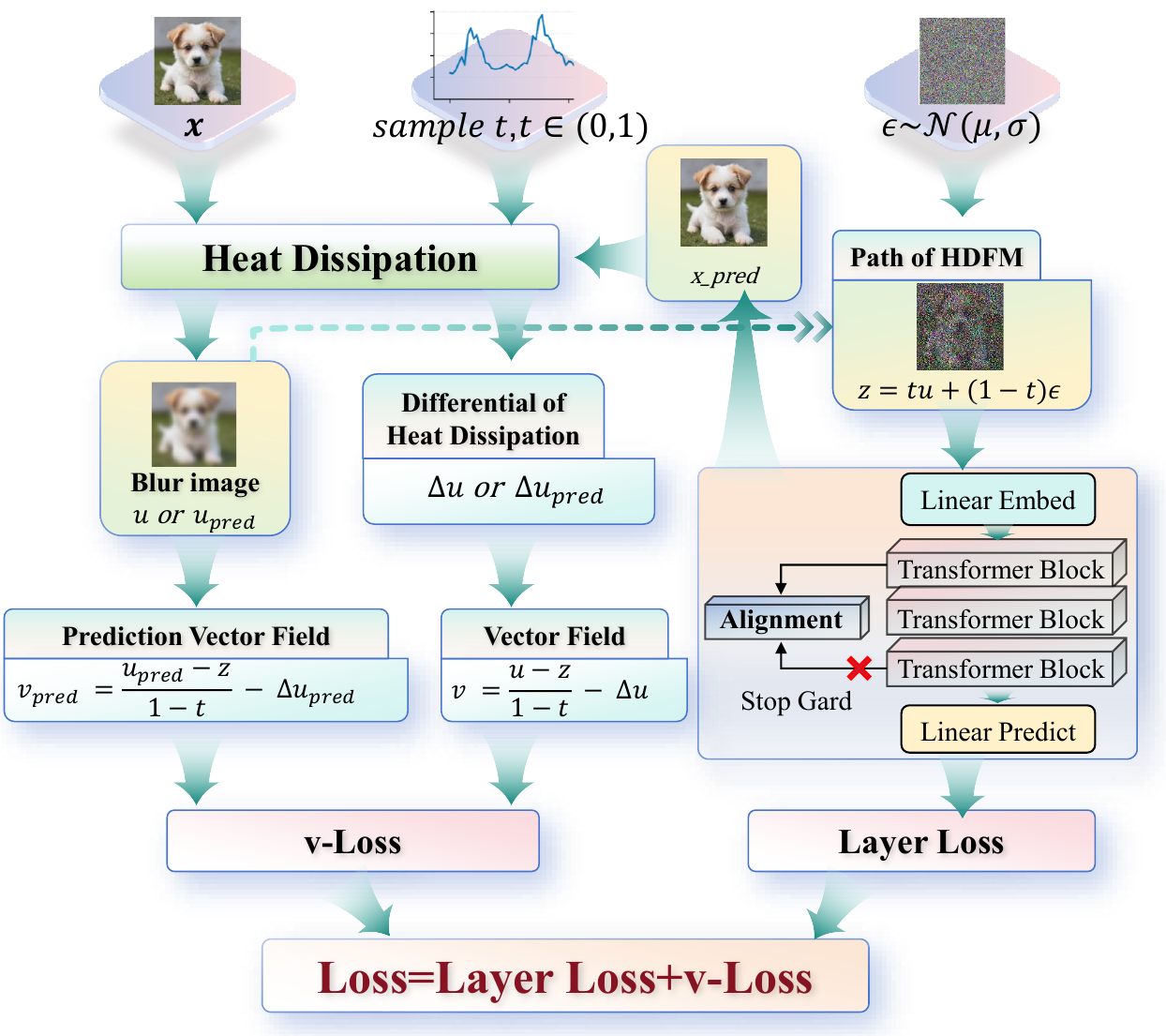}
    \caption{\textbf{The Architecture of HDFM.} The learned model can be viewed as a stack of ViT~\citep{dosovitskiy2021an} blocks, augmented with a lightweight LayerSync regularization constraint.}
    \label{fig:2}
\end{figure}

\subsection{\texorpdfstring{$x$}{x}-prediction}
JiT~\cite{li2025jit} points out that in high-dimensional observation settings, $\epsilon$-pred and $v$-pred often suffer from pronounced high-dimensional regression difficulty, whereas $x$-pred is typically more stable and effective. Therefore, JiT converts the $v$ prediction in FM into 
\begin{equation}
\label{jitv}
v=\frac{x-z_t}{1-t},
\end{equation}
shifting the prediction target from noise-based $v$-pred to $x$-pred. This can be explained by the data manifold hypothesis: real data $x$ are closer to a low-dimensional manifold, while $\epsilon$ and $v$ lie farther in the high-dimensional space, amplifying regression difficulty and error accumulation.

\subsection{Heat-dissipation Diffusion Model}
IHDM~\cite{rissanen2023generative} defines a multi-scale degradation process via the heat equation with Neumann boundary conditions:
\begin{equation}
\frac{\partial}{\partial t}u(x,y,t)=\Delta u(x,y,t).
\end{equation}
As $t\to\infty$, the image is progressively blurred to the mean. The data contracts toward a low-dimensional subspace.
Under Neumann conditions, this process can be written as
\begin{equation}
u(t)=V\exp(\Lambda t)V^\top x\quad \Leftrightarrow\quad \widetilde{u}(t)=\exp(\Lambda t)\widetilde{x},
\end{equation}
where $\Lambda$ contains negative squared-frequency terms. High-frequency components are corrupted exponentially, explicitly constructing an evolution path from coarse to fine. But it also leads to an ill-posed problem.

\subsection{Motivation of HDFM}
In FM, all multi-scale features of the data corrupt linearly with time in the forward process, preventing explicit separation of information at different scales. A commonly observed phenomenon is that generative models self-organize a coarse-to-fine path as training proceeds~\cite{rahaman2019spectral,zhi2020frequency}, suggesting a continuous cross-scale path as a favorable implicit route. However, this path is neither directly observable nor supervisable, so errors at some scale can be amplified.

IHDM explicitly models and supervises a continuous cross-scale path via heat dissipation, but it is SDE-based and unavailable to FM-style ODEs; moreover, heat-dissipation reduction still faces the curse of dimensionality by JiT theory. We propose an $x$-pred ODE method, HDFM, which addresses these issues and mitigates the ill-posed IHDM.

\section{Heat Dissipation Flow Matching (HDFM)}

Given data samples $x\sim p_{\text{data}}$ and noise $\epsilon\sim\mathcal{N}(0,I)$, HDFM learns a conditional velocity field $v_\theta(z,t,y)$ and defines the probability-flow ODE:
\begin{equation}
\frac{dz}{dt}=v_{\theta}(z,t,y),\qquad z(0)\sim\mathcal{N}(0,\sigma^2 I).
\end{equation}
Here $y$ is the class condition and $\sigma$ is the noise scale. To avoid endpoint singularities as $t\to 1$, we uniformly use $s(t)=\max(1-t,\varepsilon)$ to replace every denominator $(1-t)$. 

In our implementation, although the forward degradation in IHDM can be formulated as a spatial PDE in the continuous domain, after discretizing on the pixel grid and diagonalizing in the frequency domain, the evolution of the global image state can be equivalently written as a high-dimensional linear ODE. Therefore, this process can be naturally incorporated into an ODE-based framework.~(The proof in \textit{From a Spatial PDE to an ODE Flow}).

\subsection{HDFM Path}
\begin{figure}[t]
    \centering
    \includegraphics[width=0.48\textwidth]{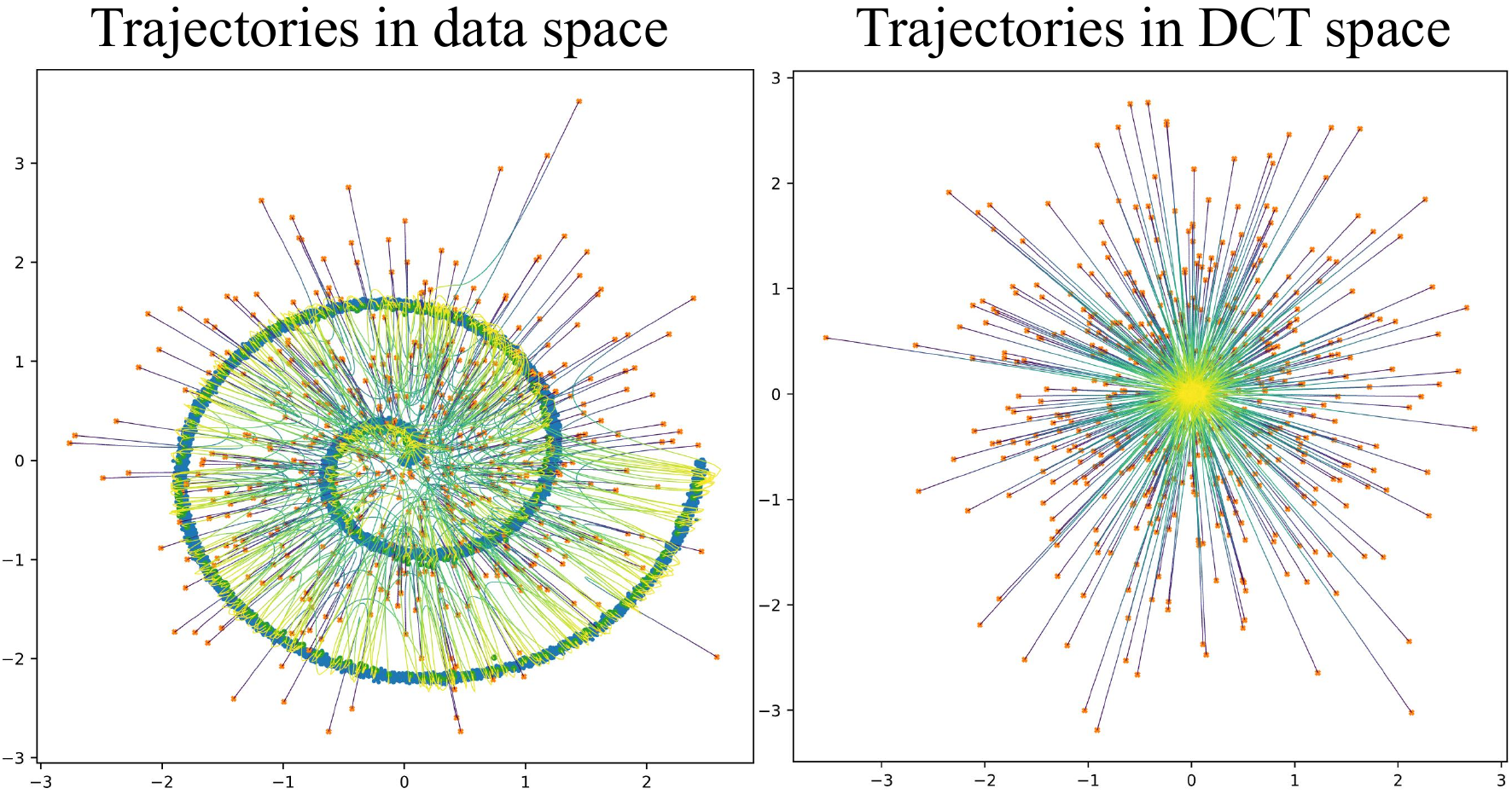}
    \caption{\textbf{HDFM transport trajectories in a toy experiment}. Left: trajectories of particles in the data space, exhibiting curved paths. Right: trajectories of the same particles in the DCT (frequency) domain, which become approximately straight lines, consistent with the linear velocity property in the frequency domain (the proof see \emph{Linear velocity in the frequency domain}).}

    \label{fig:traj}
\end{figure}

Let $\mathcal{H}_\tau$ denote the forward heat-dissipation operator, where $\tau\ge 0$ is the heat time. HDFM uses an exponential time calibration to map the flow time $t\in(0,1]$ to the heat time:
\begin{equation}
t=\exp(-\tau)\quad\Longleftrightarrow\quad \tau(t)=-\log t.
\end{equation}
Accordingly, we define the heat endpoint that varies with $t$ as
\begin{equation}
u_t \triangleq \mathcal H_{\tau(t)}(x),
\end{equation}
and construct an interpolation path that connects the noise prior and the data distribution:
\begin{equation}
\label{eq4}
z_t = t\,u_t + (1-t)\,e,\qquad e\sim\mathcal N(0,\sigma^2I).
\end{equation}
Since $t\to 0$ corresponds to $\tau\to\infty$ (strong blur) and $t\to 1$ corresponds to $\tau\to 0$ (back to the sharp image), this path explicitly injects a continuous coarse-to-fine probability path. Moreover, as Fig~\ref{fig:traj} shows, the path corresponds to a constant-velocity linear transport trajectory in DCT coordinates (the proof in \textit{Linear velocity in the frequency domain}), providing a structural explanation for HDFM's path design. We also prove that this path satisfies the continuity equation under both the marginal distribution and the conditional distribution (the proof in \textit{Continuity equation}).

\subsection{HDFM Velocity Field}
Since HDFM adopts an $x$-pred formulation~\cite{li2025jit} to directly predict the blurred data $u_t$, differentiating Eq.~(\ref{eq4}) yields
\begin{equation}
\dot z_t=\frac{u_t-z_t}{1-t}+t\,\dot u_t.
\end{equation}
We compute $\dot u_t$ by the chain rule:
\begin{equation}
\dot u_t=\frac{d}{dt}\mathcal H_{\tau(t)}(x)
=\frac{d}{d\tau}\mathcal H_{\tau}(x)\Big|_{\tau=\tau(t)}\cdot \frac{d\tau(t)}{dt}.
\end{equation}
From IHDM~\cite{rissanen2023generative}, under the Nyquist-limit setting the heat equation satisfies $\frac{d}{d\tau}\mathcal H_\tau(x)=\Delta\,\mathcal H_\tau(x)$, and $\frac{d\tau}{dt}=-\frac{1}{t}$, hence
\begin{equation}
\dot u_t=-\frac{1}{t}\Delta u_t \quad\Longrightarrow\quad t\,\dot u_t=-\Delta u_t.
\end{equation}
Substituting back, the true velocity along this path is
\begin{equation}
\label{eq8}
\dot z_t=\frac{u_t-z_t}{1-t}-\Delta u_t.
\end{equation}
Replacing $(1-t)$ by $s(t)$ and decomposing terms, we define
\begin{equation}
v_{\text{base}}(z_t,t)\triangleq \frac{u_t-z_t}{s(t)},\qquad \delta_t\triangleq \Delta u_t,
\end{equation}
so the consistent target velocity is
\begin{equation}
v^*(z_t,t)=v_{\text{base}}(z_t,t)-\delta_t
=\frac{u_t-z_t}{s(t)}-\Delta u_t.
\end{equation}
It is worth noting that ignoring $\delta_t$ is equivalent to treating a time-dependent endpoint as a fixed endpoint, which introduces systematic bias and breaks training--sampling consistency. 

\subsection{Heat-dissipation Operator}
Under Neumann boundary conditions, the discrete Laplacian operator can be diagonalized by the discrete cosine transform (DCT). Let $\tilde x=\mathrm{DCT}(x)$ and let $\Lambda$ be the diagonal matrix of Laplacian eigenvalues. Then
\begin{equation}
\tilde u(\tau)=\exp(\Lambda\tau)\tilde x,\qquad
u(\tau)=\mathrm{IDCT}\big(\tilde u(\tau)\big).
\end{equation}
With $\tau(t)=-\log t$, we have
\begin{equation}
\label{eq17}
\begin{aligned}
\tilde u_t
&=\exp(\Lambda\tau(t))\tilde x
=\exp(\Lambda(-\log t))\tilde x
=t^{-\Lambda}\tilde x,\\
u_t&=\mathrm{IDCT}(\tilde u_t).
\end{aligned}
\end{equation}
Also,
\begin{equation}
\Delta u_t=\mathrm{IDCT}\big(\Lambda\,\tilde u_t\big),
\end{equation}
so one DCT/IDCT together with pointwise multiplication in the frequency domain can obtain both $u_t$ and $\Delta u_t$, with complexity $O(N\log N)$. For the implementation of the heat-dissipation operator, see \textit{Algorithm 2}.

\subsection{Training Objective}
With $x$-pred, as shown in Fig~\ref{fig:2}, the network outputs $\hat x_\theta(z_t,t,y)$, and then the heat-dissipation operator maps it to $\hat u_t=\mathcal H_{\tau(t)}(\hat x_\theta)$ and $\Delta \hat u_t$. The final velocity prediction is defined as
\begin{equation}
v_\theta(z_t,t,y)=\frac{\hat u_t-z_t}{s(t)}-\Delta \hat u_t.
\end{equation}
During training, we construct $(u_t,\Delta u_t)$ from the ground-truth sample $x$ and form the consistent supervision $v^*$, minimizing the velocity regression loss:
\begin{equation}
\label{eq15}
\mathcal{L}_{\text{vel}}=\mathbb{E}\Big[\big\|v_\theta(z_t,t,y)-v^*(z_t,t)\big\|_2^2\Big].
\end{equation}
In addition, to enhance representation consistency, we introduce a lightweight LayerSync~\cite{haghighi2025layersync} regularizer $\mathcal L_{\text{LS}}$: it aligns patch features of selected weak/strong layers via cosine alignment and applies stop-grad to the strong branch. The total loss is
\begin{equation}
\mathcal{L}=\mathcal{L}_{\text{vel}}+\lambda\,\mathcal{L}_{\text{LS}}.
\end{equation}
The train step implementation can see \textit{Algorithm 1}.

\subsection{Sample}
At sampling process, we compute the velocity as
\begin{equation}
\label{equ:17}
\hat v_{base}=\frac{\hat u_t-z_t}{s(t)},\\
\hat\delta=\Delta \hat u_t,\\
v_{\theta}=\hat v_{base}-\hat\delta,
\end{equation}
and integrate it with standard ODE solvers (\textit{e.g.}, Euler/Heun). About the implementation of the sampling process can see \textit{ Algorithm 3}.

\noindent\textbf{Interval CFG.}
To maintain consistency of the computation structure, we compute $\hat v_{base}$ and $\hat\delta$ on unconditional/conditional branches and combine them with CFG. The final velocity is
\begin{equation}
\label{equ:18}
\left\{
\begin{aligned}
\hat v_\text{base}^{\text{cfg}}&=\hat v_\text{base}^u + \alpha\big(\hat v_\text{base}^c-\hat v_\text{base}^u\big),\\
\hat\delta^{\text{cfg}}&=\hat\delta^u + \alpha\big(\hat\delta^c-\hat\delta^u\big),\\
v_{\text{cfg}}&=\hat v_\text{base}^{\text{cfg}}-\hat\delta^{\text{cfg}}.
\end{aligned}
\right.
\end{equation}
We set $\alpha > 1$ only when $t \in (t_{\min}, t_{\max})$, where $t_{\min}$ and $t_{\max}$ are both hyper-parameters. Otherwise, we set $\alpha = 1$ to improve numerical stability and avoid guidance-induced detail degradation. Here, $\hat v_\text{base}^c$ denotes the base component obtained under the conditional branch by 
substituting $\hat u_c$ into Eq.~(\ref{equ:17}), and $\hat\delta^c$ is defined analogously. Similarly, $\hat v_\text{base}^u$ and $\hat\delta^u$ denote the CFG velocity components under the unconditional branch.

\noindent\textbf{Adaptive correction Fusion.}
Under finite-step integration and model approximation error, the full correction is not stepwise optimal. We introduce $\beta\in[0.05,0.95]$ and use
\begin{equation}
v=v_\text{base}-\beta\,\delta,
\end{equation}
where $\beta$ is estimated from the energy of a local one-step finite-difference residual: we construct residual energies $\sigma_{\text{full}}$ and $\sigma_{\text{no}}$ using the changes between adjacent time points under the full velocity ($v_{base}-\delta$) and the no-delta velocity ($v_{base}$), respectively, and set
\begin{equation}
\beta=\mathrm{clip}\!\left(\frac{\sigma_{\text{no}}}{\sigma_{\text{full}}+\sigma_{\text{no}}},\,0.05,\,0.95\right).
\end{equation}
This strategy adaptively adjusts the correction strength at different stages, improving stability and quality for few-step sampling.

\subsection{About Ill-posed Challenge}
The ill-posed challenge of IHDM is rooted in the exponential blurring of its forward process: once the inverse dynamics are learned, any perturbation of the mean (\textit{e.g}., observation noise or limited numerical precision) can be exponentially amplified. HDFM reparameterizes time, converting exponential blurring into a power law and thereby fundamentally weakening this amplification.

This power-law blurring introduces a near-$t=0$ singular corruption behavior—visually, blur changes more abruptly as $t\to0$. However, HDFM avoids the ill-posed challenge because its flow takes the form $z=tu_t+(1-t)e$. As $t\to 0$, even if $u_t$ varies sharply, the factor $t$ suppresses its impact so that $tu_t\to0$ and $z\approx e$. Hence, compared with IHDM, HDFM substantially mitigates the intrinsic ill-posed challenge.

\section{Experiment}
\subsection{Setup}
The compared methods fall into two categories. For heat-dissipation-related approaches, we choose IHDM~\cite{rissanen2023generative} and Blurring~\cite{hoogeboom2023blurring} as the main baselines: IHDM represents the classical inverse heat-dissipation generative framework, while Blurring is a recent heat-dissipation diffusion model with strong empirical performance. For non-heat-dissipation methods, we include the latest JiT~\cite{li2025jit} as a baseline. JiT exhibits strong overall performance within the FM family and also serves as the backbone architecture of our method, enabling us to verify the gains brought by introducing the heat-dissipation path and consistent supervision under an identical backbone. 

Since this paper is not intended to study specific model architectures, we adopt the publicly released JiT code and its default B/16 configuration as the backbone, and report JiT’s performance under the same setting for comparison. All methods are downloaded from their official public repositories, and all hyperparameters follow the default configurations of the released code.

We evaluate all methods on three public image datasets: ImageNet256~\cite{imagenet15russakovsky}, Places365-small~\cite{zhou2017places}, and ImageNet128~\cite{imagenet15russakovsky}. In addition to the main comparisons, we conduct systematic ablation studies to validate the contribution and necessity of each key component of our method.

\begin{figure}
    \centering
    \includegraphics[width=0.5\textwidth]{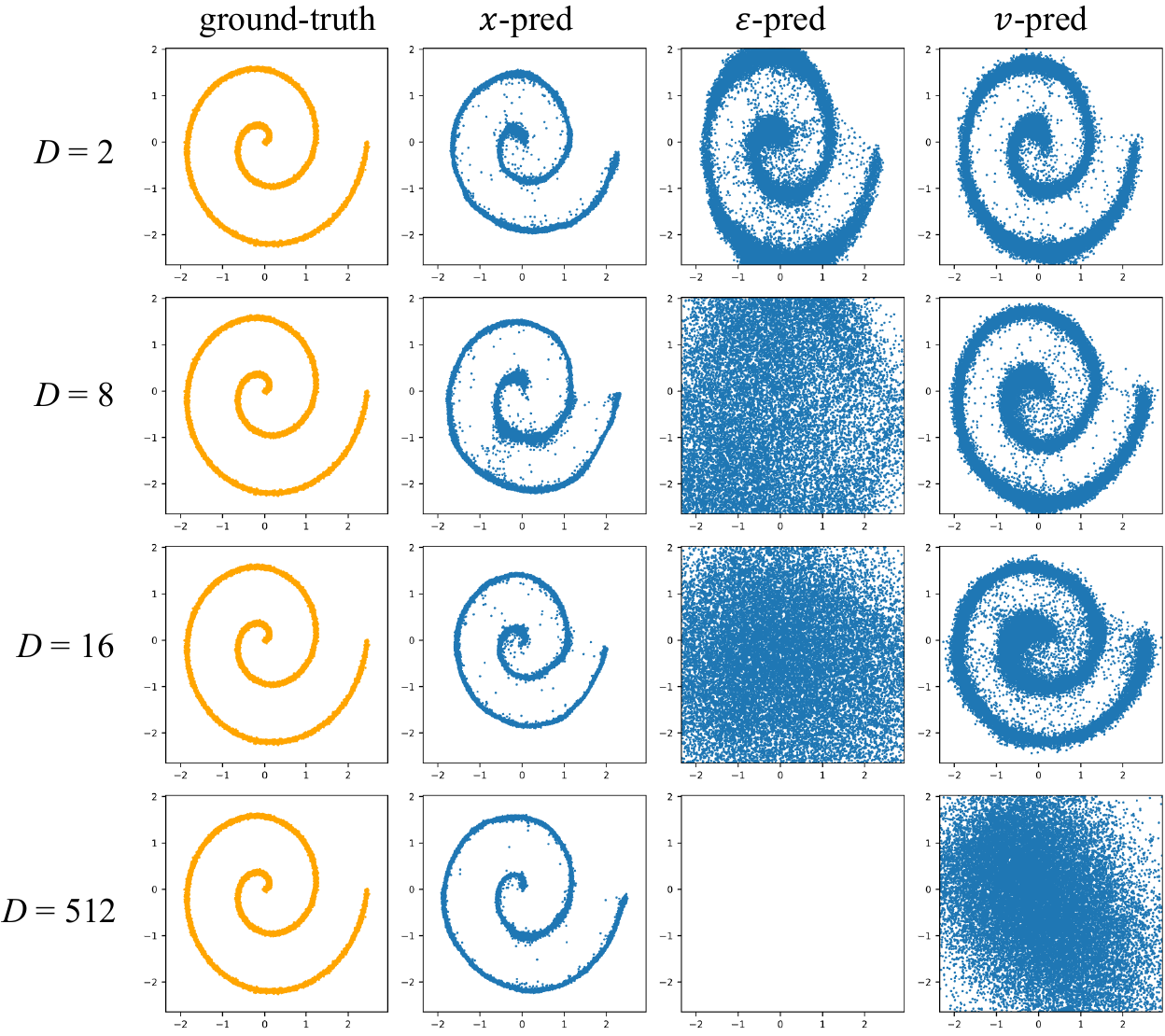}
    \caption{\textbf{Toy Experiment in Heat Dissipation}: a 2D dataset is “buried” into a D-dimensional space via a fixed orthogonal projection matrix. After applying 1D heat-dissipation blurring, a lightweight neural network is trained to reconstruct the data during inverse heat dissipation, and we visualize the resulting.}
    \label{fig:toy}
\end{figure}

\subsection{Toy Experiment}

To verify whether the assumption of data-manifold~\cite{chapelle2006b,li2025jit} still holds under heat dissipation, we construct a controlled setting where a $d$-dimensional underlying distribution is ``buried'' in a higher-dimensional observed space ($d \ll D$).
Concretely, we sample a 2D spiral $\hat{x}\in\mathbb{R}^{2}$ and embed it into $\mathbb{R}^{D}$ by a fixed random \emph{column-orthogonal} projection matrix $P\in\mathbb{R}^{D\times 2}$ satisfying $P^{\top}P=I$:
\begin{equation}
    x = P \hat{x}\in\mathbb{R}^{D}.
\end{equation}
The matrix $P$ is unknown to the model. It can also be used for visualization by transforming generated $D$-dimensional samples back to the 2D plane via $\hat{x}=P^{\top}x$.
We consider $D\in\{2,8,16,512\}$ while keeping the underlying dimension fixed at $d=2$.

This toy experiment adopts a simplified inverse-heat forward corruption along the feature dimension, implemented via a 1D DCT heat kernel.
Given $x_0$, we define a blurred state $u_t$ and its Laplacian term $\Delta u_t$, and substitute them into Eq.~(\ref{eq4}) and Eq.~(\ref{eq8}) to obtain $z_t$ and the corresponding velocity $v_t$. Meanwhile, we keep the same $v$-loss for all parameterizations as Eq.~(\ref{eq15}) and only change what the network predicts.

\paragraph{Parameterizations.}
We train a fixed-capacity 5-layer ReLU MLP conditioned on $t$, while varying the network output space:
(i) $x$-prediction: the network predicts $\hat{x}_0$ (clean sample) which is then mapped to $v_{\theta}$ by applying the Eq.~(\ref{eq17}) and plugging into the Eq.~(\ref{eq8});
(ii) $v$-prediction: the network directly predicts $v_{\theta}$;
(and we also include $\epsilon$-prediction for completeness).

\paragraph{Observation.}
Fig.~\ref{fig:toy} visualizes the generated samples projected back to the 2D plane.
When $D=2$, both $x$-pred and $v$-pred can capture the spiral distribution.
However, as $D$ increases, the $v$-pred outputs become progressively \emph{noisier} and less structured:
already at moderate $D$ the generated samples thicken into a diffuse cloud, and at very large $D$ the generation collapses to an almost unstructured scatter.
In contrast, $x$-prediction remains visually faithful to the underlying spiral even when the ambient dimension is large, producing substantially cleaner samples with far less off-manifold noise.

\paragraph{Why does $x$-prediction scale better with $D$?}
In heat-dissipation construction, the true data distribution occupies (approximately) only the 2D subspace spanned by $P$.
Thus, the \emph{ideal} output is intrinsically low-dimensional even though it is represented in $\mathbb{R}^{D}$.
An $x$-predictor can learn to recover this structured, low-dimensional signal while implicitly ignoring irrelevant directions orthogonal to the data subspace.
By contrast, $v$ (and similarly $\epsilon$) is an off-manifold target whose energy is distributed across the ambient coordinates.
As $D$ grows under a fixed-capacity network, accurately regressing such dense off-manifold quantities becomes increasingly difficult; the residual errors manifest as isotropic noise and are further amplified through multi-step ODE integration during sampling.
This toy result supports our design choice of emphasizing $x$-prediction in high-dimensional settings: it preserves the low-dimensional structure and yields markedly less noisy generation as the ambient dimension increases.

\begin{figure}[t]
    \centering
    \includegraphics[width=0.48\textwidth]{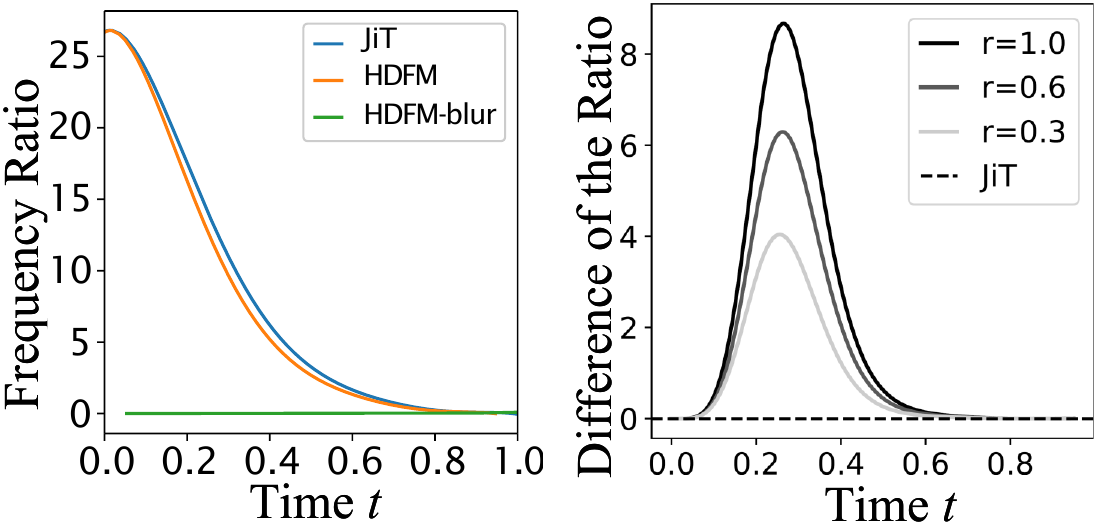}
    \caption{\textbf{Spectral discrepancy during sampling.} Left: Frequency ratio curves for JiT (full noise), HDFM, and HDFM-blur (pure blurring), where HDFM exhibits an intermediate behavior between the two extremes.  Right: Attenuating the blur strength in HDFM by a factor $r$ shifts the curve toward JiT, validating a smooth interpolation from blur-dominant to noise-dominant sampling. These suggest that HDFM can get a hybrid ability to recover low-frequency structure from noise and high-frequency details from blur.}
    \label{fig:spec}
\end{figure}

\paragraph{Image-space validation of $x$-pred vs.\ $v$-pred.}
To further examine the parameterization gap between $v$-prediction and $x$-prediction under the HDFM framework, we conduct experiments on real image datasets, including ImageNet256, ImageNet128, and Places365.
Both variants share the \emph{same} network backbone and training process, and differ only in the prediction target: $x$-pred regresses an endpoint variable aligned with the data manifold, whereas $v$-pred directly regresses the velocity field. Across all datasets, $x$-pred consistently achieves substantially better generation quality than $v$-pred (see Table~\ref{tab:1}). These observations corroborate our toy findings: in high-dimensional image spaces, directly regressing $v$ (or $\epsilon$) constitutes a dense off-manifold regression problem that is difficult for limited-capacity networks to learn reliably.

These suggest that $x$-prediction aligns the learning target with the heat-dissipation data distribution to deterministic computations, leading to markedly improved stability and sample quality.
Taken together, these results motivate our default choice of $x$-prediction in HDFM.

\subsection{Quantitative Experiments on Heat-Dissipation Blurring}

\begin{table}
\caption{FID of $v$-pred and $x$-pred HDFM on image datasets.}
\centering
\label{tab:1}
\begin{tabular}{cccc}
\toprule
         & Imagenet256 & Imagenet128 & Place365 \\
\midrule
$x$-pred & 9.40       & 12.95    & 27.16   \\  
$v$-pred & 253.40      & 109.56      & 316.50    \\  
\bottomrule
\end{tabular}
\end{table}

To quantify the impact of heat-dissipation blurring strength on HDFM, we design a quantitative study to analyze how the frequency-domain energy distribution evolves under different conditions, with particular focus on frequency-shift behavior. We define the metric as the frequency ratio $ratio=\frac{E_{\text{high}}}{E_{\text{low}}}$, where $E_{\text{high}}$ and $E_{\text{low}}$ denote the high- and low-frequency energies, respectively. 


The experiment consists of two parts as Fig~\ref{fig:spec}. The first compares the overall shapes of frequency-ratio curves under three basic settings: (1) the original JiT model, serving as a ``no blur, noise only'' baseline; (2) HDFM, representing the ``blur + noise'' setting; and (3) HDFM-blur~(see \textit{Algorithm 4} and \textit{Algorithm 5}), an extreme ``blur only, no noise'' setting. The second part further compares the frequency-ratio difference curves between HDFM and JiT. Taking JiT as the reference, we decrease the blur-strength parameter $r$ from 1.0 (HDFM) to 0.3 to examine whether HDFM approaches JiT as blurring weakens. Convergence would indicate that HDFM indeed benefits from the introduced blurring compared with JiT. All curves are obtained by averaging over a shared time grid after randomly sampling 5000 images.

As shown in the left plot of Fig.~\ref{fig:spec}, HDFM and JiT share the same starting point because both are initialized from full-band high-frequency noise; in contrast, HDFM-blur starts from a strongly blurred image dominated by low frequencies and progressively restores high frequencies, hence its ratio begins near zero. The three methods converge to a similar endpoint ($\sim 0.075$), consistent with the empirical statistics of natural images where low-frequency energy dominates high-frequency energy~\cite{field1987relations}. Throughout sampling, HDFM remains between JiT and HDFM-blur, while exhibiting a faster decreasing trend. 

The blur-strength decrease experiment~(the right plot of Fig.~\ref{fig:spec}) further supports this observation: as the blurring strength is gradually reduced (from $1$ to $0.3$), the frequency-ratio curve consistently approaches the JiT baseline. This indicates that blurring strength is a key factor shaping HDFM's frequency-domain transport behavior and the associated velocity field.

These suggest that HDFM can get a hybrid ability to recover low-frequency structure from noise and high-frequency details from blur. The hybrid ability may accelerate convergence of the spectral energy allocation toward natural-image statistics.

\begin{figure}[t]
    \centering
    \includegraphics[width=0.48\textwidth]{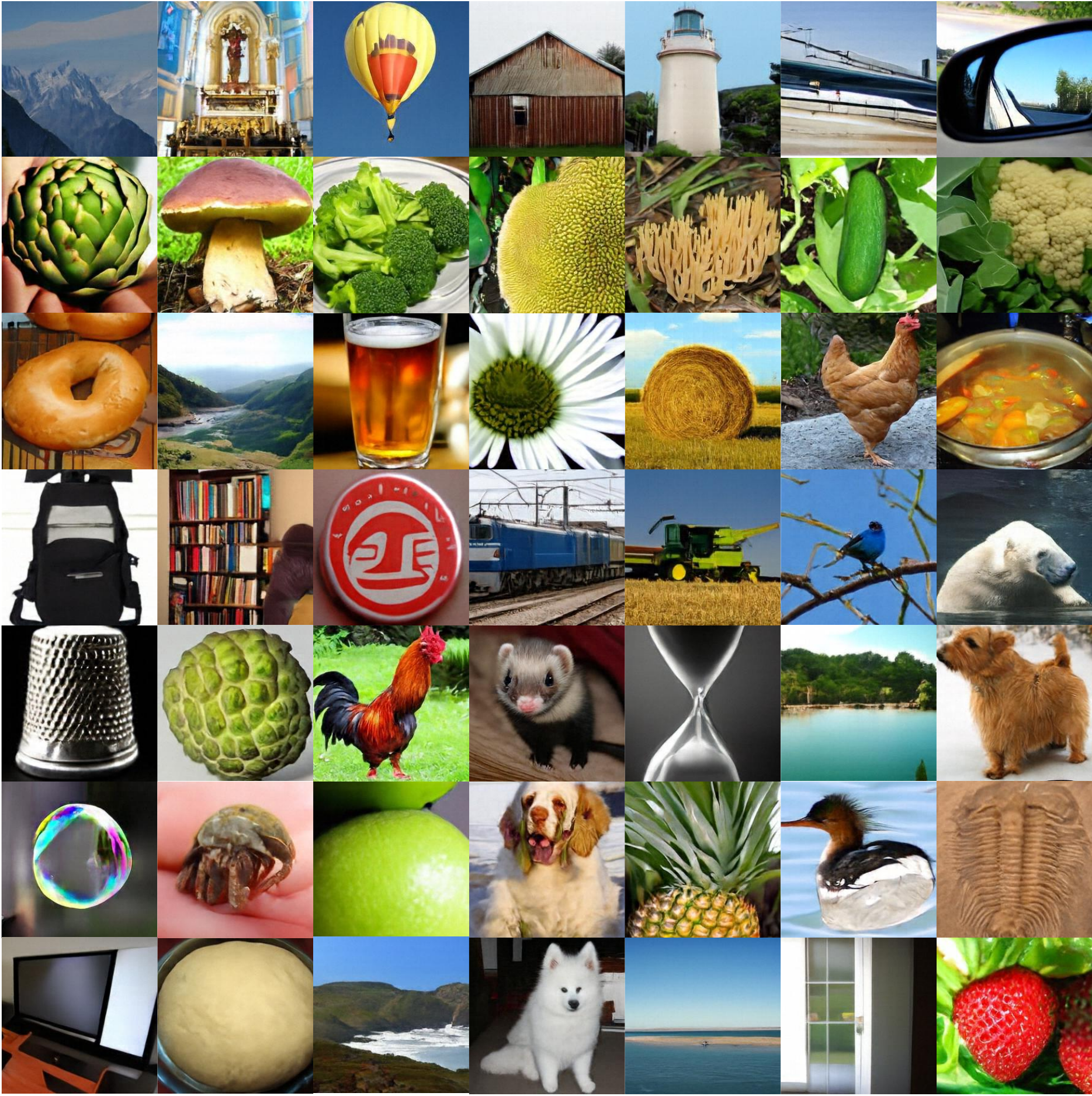}
    \caption{\textbf{Qualitative Results.} Some examples are selected on ImageNet256.}
    \label{fig:sample}
\end{figure}

\subsection{Comparisons}

\begin{table}[t]
  \centering
  \small
  \caption{Quantitative comparison across datasets.}
  \label{tab:main_results}
  \setlength{\tabcolsep}{3pt}
  \renewcommand{\arraystretch}{1.12}

  \begin{tabular}{l l c}
    \toprule
    Dataset & Method & FID$\downarrow$ \\
    \midrule
    \multirow{4}{*}{ImageNet256}
      & IHDM~\cite{rissanen2023generative}       & 70.84 \\
      & Blurring~\cite{hoogeboom2023blurring}    & 36.57 \\
      & $f$-DM-DS~\cite{gu2023fdm}  &  8.2\\
      & JiT-B/16~\cite{li2025jit}                & 8.17 \\
     & HDFM (ours)                              & 7.73  \\
    \midrule

    \multirow{4}{*}{ImageNet128}
      & IHDM~\cite{rissanen2023generative}        & 56.51 \\
     & Blurring~\cite{hoogeboom2023blurring}    & 28.92 \\
     & JiT-B/16~\cite{li2025jit}                & 18.47 \\
       & HDFM (ours)                              & 12.92 \\
    \midrule

    \multirow{4}{*}{Places365}
     & IHDM~\cite{rissanen2023generative}        & 78.11 \\
      & Blurring~\cite{hoogeboom2023blurring}     & 42.37 \\
      & JiT-B/16~\cite{li2025jit}                & 29.60 \\
      & HDFM (ours)                              & 27.16 \\
    \bottomrule
  \end{tabular}
\end{table}

As shown in the Table~\ref{tab:main_results}, we compare blur-based methods and JiT on ImageNet256, ImageNet128, and Places365~(all experimental settings can see \textit{Comparison Methods Setup}). Across all three datasets, HDFM achieves the best or tied-best overall performance. On ImageNet256, HDFM achieves more consistent improvements in structural coherence and fine-detail fidelity at high resolutions. Results on ImageNet128 suggest that the designed path/dynamics remains transferable and robust at lower resolutions. On the more scene-diverse Places365 dataset, HDFM demonstrates stable modeling capability in handling complex semantic distributions.

To complement the quantitative results, we provide sampling visualizations on ImageNet256 in Fig.~\ref{fig:sample}. Qualitatively, HDFM produces more natural samples with clearer object contours, stronger semantic consistency, and richer local textures, while exhibiting fewer edge and texture artifacts.

\subsection{Parameter Analysis}


\begin{table}[t]
  \centering

  \begin{minipage}[t]{0.48\columnwidth}
    \centering
    \captionof{table}{Effect of CFG.}
    \label{tab:cfg_ablation}
    \small
    \setlength{\tabcolsep}{4pt}
    \renewcommand{\arraystretch}{1.1}
    \begin{tabular}{lcc}
      \toprule
      CFG & IS$\uparrow$ & FID$\downarrow$ \\
      \midrule
      0.0 & 14.83  & 72.58 \\
      2.9 & 149.53 & 10.46 \\
      3.5 & 173.96 & 9.40 \\
      \bottomrule
    \end{tabular}
  \end{minipage}
  \hfill
  \begin{minipage}[t]{0.48\columnwidth}
    \centering
    \captionof{table}{Effect of Noise scale.}
    \label{tab:noise_ablation}
    \small
    \setlength{\tabcolsep}{4pt}
    \renewcommand{\arraystretch}{1.1}
    \begin{tabular}{lcc}
      \toprule
      Noise scale & IS$\uparrow$ & FID$\downarrow$ \\
      \midrule
      1.0 & 218.50 & 7.73 \\
      0.8 & 184.03 & 9.27 \\
      0.5 & 139.17 & 11.90 \\
      \bottomrule
    \end{tabular}
  \end{minipage}

\end{table}

This section focuses on two hyperparameters that most strongly affect performance: the CFG scale and the noise scale, which control conditioning strength and generative capacity, respectively. The following two subsections discuss them in detail.

\subsubsection{CFG Scale}
As shown by the ImageNet256 hyper-parameter study in Table~\ref{tab:cfg_ablation}, the CFG scale has a noticeable impact on conditional generation quality. Without guidance (CFG$=0.0$), the model lacks explicit class conditioning, leading to poor semantic alignment and low recognizability. Increasing the guidance to CFG$=2.9$ strengthens conditional information and substantially improves the match to the data distribution. Further increasing CFG to 3.5 yields the best overall performance, indicating that stronger guidance within this range continues to improve semantic alignment and fine-detail quality. 

\subsubsection{Noise Scale}


From the ImageNet256 ablation in Table \ref{tab:noise_ablation}, the noise scale has a pronounced and monotonic impact on generation quality and diversity. Increasing the noise scale from 0.5 to 0.8 and then to 1.0 consistently raises IS and lowers FID. This indicates that, under our setting, stronger stochasticity improves sampling coverage and distribution alignment, mitigating mode collapse and poor local minima, and thus enhancing both semantic fidelity and realism. In contrast, weaker noise (\textit{e.g.}, 0.5) makes sampling more deterministic, limiting diversity and detail recovery, which results in lower IS and higher FID.

\section{Related Work}
Diffusion-based image generation models~\cite{sohl2015deep} and multi-stage generative frameworks such as Flow Matching~\cite{lipman2023flow,liu2023flow,albergo2023building} have achieved remarkable success across a wide range of domains~\cite{lewis2025scalable,he2025diffusion}. Most of these models, however, rely on forward and reverse processes driven by isotropic Gaussian noise. In recent years, alternative image corruption processes have been explored, offering diffusion dynamics with different properties and inductive biases. In particular, the inverse heat dissipation (IHD) process~\cite{rissanen2023generative} proposed uses a heat-based blurring equation with a small amount of noise as the forward process, providing an explicit coarse-to-fine structural prior. The Blurring model~\cite{hoogeboom2023blurring} further extends this line of work, scheduling the blur process together with Gaussian noise to achieve improved performance. These approaches explicitly inject observation noise to avoid making the reverse process fully deterministic. By contrast, cold diffusion~\cite{bansal2023cold} proposes a fully deterministic, noise-free reverse sampling procedure, training the model to learn reverse dynamics beyond strict PDE formulations. In another direction, Beyond Blur~\cite{gruszczynski2025beyond} introduces stochastic velocity fields and proposes a turbulence-inspired formulation. Ours work adopts a noise-injected inverse heat path, motivated by the observation that once the ill-posedness of inverse heat dissipation is addressed, the resulting process can be naturally integrated in the frequency domain and complement conventional approaches that rely on recovering images from high-frequency noise.

The manifold assumption for natural data has a long history~\cite{roweis2000nonlinear,tenenbaum2000global}, with extensive classical studies spanning manifold learning and generative modeling~\cite{loaizadeep,chendeconstructing}. Among them, several works explicitly connect generative modeling to manifold structure. JiT~\cite{li2025jit} observes that as data become higher-dimensional, reconstructing data from noise becomes less accurate, and propose $x$-prediction to mitigate the mismatch between data residing on a low-dimensional manifold and learning targets in a high-dimensional noise space. While not the earliest work advocating this idea~\cite{ho2020denoising,SiDEmiel}, their experiments—especially under settings where data are embedded into higher-dimensional spaces—provide empirical evidence for its effectiveness. In our setting, we observe that blurring-based processes face a similar, and often more severe, mismatch. Accordingly, we adopt $x$-pred as a primary modeling choice to alleviate this dimensional misalignment.

\section{Conclusion and Discussion}

We propose HDFM, a continuous cross-scale path algorithm for coarse-to-fine generation. As a principled framework, it injects the multiscale structure of heat dissipation into flow matching. From a frequency-transport perspective, HDFM balances the noise-driven and blur-driven generation routes.

HDFM makes three core contributions: (1) it provides flow matching with a new explicit continuous cross-scale path; (2) by reparameterizing time and interpolating along the heat-dissipation path, it alleviates the ill-posed of inverse heat dissipation; and (3) with $x$-prediction, it mitigates the high-dimensional regression difficulty implied by the data-manifold assumption. Toy experiments verify both the high-dimensional regression challenge under heat dissipation and the effectiveness of our remedy, and further demonstrate the path’s efficient linear transport in the DCT frequency domain. We conduct quantitative studies over blur strength, showing that blurring can benefit purely noise-based methods. Finally, HDFM outperforms the baseline on three datasets, and we analyze the key hyperparameters, paving the way for extending the approach to broader applications.

\section*{Ethical Statement}

There are no ethical issues.

\bibliographystyle{named}
\bibliography{ijcai26}

@article{watson2023novo,
  title={De novo design of protein structure and function with RFdiffusion},
  author={Watson, Joseph L and Juergens, David and Bennett, Nathaniel R and Trippe, Brian L and Yim, Jason and Eisenach, Helen E and Ahern, Woody and Borst, Andrew J and Ragotte, Robert J and Milles, Lukas F and others},
  journal={Nature},
  volume={620},
  number={7976},
  pages={1089--1100},
  year={2023},
  publisher={Nature Publishing Group UK London}
}

@article{bjorck2025gr00t,
  title={Gr00t n1: An open foundation model for generalist humanoid robots},
  author={Bjorck, Johan and Casta{\~n}eda, Fernando and Cherniadev, Nikita and Da, Xingye and Ding, Runyu and Fan, Linxi and Fang, Yu and Fox, Dieter and Hu, Fengyuan and Huang, Spencer and others},
  journal={arXiv preprint arXiv:2503.14734},
  year={2025}
}

@inproceedings{sohl2015deep,
  title={Deep unsupervised learning using nonequilibrium thermodynamics},
  author={Sohl-Dickstein, Jascha and Weiss, Eric and Maheswaranathan, Niru and Ganguli, Surya},
  booktitle={ICML},
  year={2015},
}

@inproceedings{lipman2023flow,
  title={Flow Matching for Generative Modeling},
  author={Lipman, Yaron and Chen, Ricky TQ and Ben-Hamu, Heli and Nickel, Maximilian and Le, Matt},
  booktitle={ICLR},
  year={2023}
}

@inproceedings{liu2023flow,
  title={Flow Straight and Fast: Learning to Generate and Transfer Data with Rectified Flow},
  author={Liu, Xingchao and Gong, Chengyue and Liu, Qiang},
  booktitle={ICLR},
  year={2023}
}

@inproceedings{albergo2023building,
  title={Building Normalizing Flows with Stochastic Interpolants},
  author={Albergo, Michael and Vanden-Eijnden, Eric},
  booktitle={ICLR},
  year={2023}
}

@article{lewis2025scalable,
  title={Scalable emulation of protein equilibrium ensembles with generative deep learning},
  author={Lewis, Sarah and Hempel, Tim and Jim{\'e}nez-Luna, Jos{\'e} and Gastegger, Michael and Xie, Yu and Foong, Andrew YK and Satorras, Victor Garc{\'\i}a and Abdin, Osama and Veeling, Bastiaan S and Zaporozhets, Iryna and others},
  journal={Science},
  volume={389},
  number={6761},
  pages={eadv9817},
  year={2025},
  publisher={American Association for the Advancement of Science}
}

@article{he2025diffusion,
  title={Diffusion models in low-level vision: A survey},
  author={He, Chunming and Shen, Yuqi and Fang, Chengyu and Xiao, Fengyang and Tang, Longxiang and Zhang, Yulun and Zuo, Wangmeng and Guo, Zhenhua and Li, Xiu},
  journal={IEEE Transactions on Pattern Analysis and Machine Intelligence},
  year={2025},
  publisher={IEEE}
}

@inproceedings{rissanen2023generative,
  title={Generative Modelling with Inverse Heat Dissipation},
  author={Rissanen, Severi and Heinonen, Markus and Solin, Arno},
  booktitle={ICLR},
  year={2023}
}

@inproceedings{hoogeboom2023blurring,
  title={Blurring Diffusion Models},
  author={Hoogeboom, Emiel and Salimans, Tim},
  booktitle={ICLR},
  year={2023}
}

@inproceedings{bansal2023cold,
  title={Cold diffusion: Inverting arbitrary image transforms without noise},
  author={Bansal, Arpit and Borgnia, Eitan and Chu, Hong-Min and Li, Jie and Kazemi, Hamid and Huang, Furong and Goldblum, Micah and Geiping, Jonas and Goldstein, Tom},
  booktitle={NeurIPS},
  year={2023}
}

@inproceedings{ho2020denoising,
  title={Denoising diffusion probabilistic models},
  author={Ho, Jonathan and Jain, Ajay and Abbeel, Pieter},
  booktitle={NeurIPS},
  year={2020}
}

@inproceedings{songscore,
  title={Score-Based Generative Modeling through Stochastic Differential Equations},
  author={Song, Yang and Sohl-Dickstein, Jascha and Kingma, Diederik P and Kumar, Abhishek and Ermon, Stefano and Poole, Ben},
  booktitle={ICLR},
  year={2020}
}

@article{chapelle2006b,
  title={SemiSupervised Learning},
  author={Olivier Chapelle and Bernhard Sch\"olkopf and Alexander Zien},
  journal={MIT Press},
  year={2006}
}

@article{li2025jit,
  title={Back to Basics: Let Denoising Generative Models Denoise},
  author={Li, Tianhong and He, Kaiming},
  journal={arXiv preprint arXiv:2511.13720},
  year={2025}
}

@inproceedings{gruszczynski2025beyond,
  title={Beyond Blur: A Fluid Perspective on Generative Diffusion Models},
  author={Gruszczynski, Grzegorz and Meixner, Jakub and Wlodarczyk, Michal and Musialski, Przemyslaw},
  booktitle={ICCV},
  year={2025}
}

@article{roweis2000nonlinear,
  title={Nonlinear dimensionality reduction by locally linear embedding},
  author={Roweis, Sam T and Saul, Lawrence K},
  journal={science},
  volume={290},
  number={5500},
  pages={2323--2326},
  year={2000},
  publisher={American Association for the Advancement of Science}
}

@article{tenenbaum2000global,
  title={A global geometric framework for nonlinear dimensionality reduction},
  author={Tenenbaum, Joshua B and Silva, Vin de and Langford, John C},
  journal={science},
  volume={290},
  number={5500},
  pages={2319--2323},
  year={2000},
  publisher={American Association for the Advancement of Science}
}

@article{loaizadeep,
  title={Deep Generative Models through the Lens of the Manifold Hypothesis: A Survey and New Connections},
  author={Loaiza-Ganem, Gabriel and Ross, Brendan Leigh and Hosseinzadeh, Rasa and Caterini, Anthony L and Cresswell, Jesse C},
  journal={Transactions on Machine Learning Research},
  year={2024}
}

@inproceedings{chendeconstructing,
  title={Deconstructing Denoising Diffusion Models for Self-Supervised Learning},
  author={Chen, Xinlei and Liu, Zhuang and Xie, Saining and He, Kaiming},
  booktitle={ICLR},
  year={2025}
}

@inproceedings{SiDEmiel,
  title={Simpler Diffusion (SiD2): 1.5 FID on ImageNet512 with pixel-space diffusion.},
  author={Emiel, Hoogeboom and Thomas, Mensink and Jonathan, Heek and Kay Lamerigts and Ruiqi, Gao and Tim, Salimans},
  booktitle={CVPR},
  year={2025}
}

@inproceedings{dosovitskiy2021an,
title={An Image is Worth 16x16 Words: Transformers for Image Recognition at Scale},
author={Alexey Dosovitskiy and Lucas Beyer and Alexander Kolesnikov and Dirk Weissenborn and Xiaohua Zhai and Thomas Unterthiner and Mostafa Dehghani and Matthias Minderer and Georg Heigold and Sylvain Gelly and Jakob Uszkoreit and Neil Houlsby},
booktitle={ICLR},
year={2021}
}

@article{chen2018neural,
  title={Neural ordinary differential equations},
  author={Chen, Ricky TQ and Rubanova, Yulia and Bettencourt, Jesse and Duvenaud, David K},
  journal={NeurIPS},
  volume={31},
  year={2018}
}

@inproceedings{gu2023fdm,
    title={f-DM: A Multi-stage Diffusion Model via Progressive Signal Transformation},
    author={Gu, Jiatao and Zhai, Shuangfei and Zhang, Yizhe and Bautista, Miguel Angel and Susskind, Josh},
    booktitle={ICLR},
    year={2023}
  }

@article{rahaman2019spectral,
  title={On the spectral bias of neural networks: International Conference on Machine Learning},
  author={Rahaman, N and Baratin, A and Arpit, D and Draxler, F and Lin, M and Hamprecht, F and Bengio, Y and Courville, A},
  journal={arXiv},
  year={2019}
}

@article{zhi2020frequency,
  title={Frequency principle: Fourier analysis sheds light on deep neural networks},
  author={Zhi-Qin, John and Xu, Yaoyu Zhang and Tao, Luo and Yanyang, Xiao and Zheng, Ma},
  journal={Communications in Computational Physics},
  volume={28},
  number={5},
  pages={1746--1767},
  year={2020},
  publisher={Global Science Press}
}

@article{haghighi2025layersync,
  title={LayerSync: Self-aligning Intermediate Layers},
  author={Haghighi, Yasaman and van Delft, Bastien and Hassan, Mariam and Alahi, Alexandre},
  journal={arXiv preprint arXiv:2510.12581},
  year={2025}
}

@article{imagenet15russakovsky,
    Author = {Olga Russakovsky and Jia Deng and Hao Su and Jonathan Krause and Sanjeev Satheesh and Sean Ma and Zhiheng Huang and Andrej Karpathy and Aditya Khosla and Michael Bernstein and Alexander C. Berg and Li Fei-Fei},
    Title = { {ImageNet Large Scale Visual Recognition Challenge} },
    Year = {2015},
    journal   = {International Journal of Computer Vision (IJCV)},
    doi = {10.1007/s11263-015-0816-y},
    volume={115},
    number={3},
    pages={211-252}
}

@article{zhou2017places,
  title={Places: A 10 million Image Database for Scene Recognition},
  author={Zhou, Bolei and Lapedriza, Agata and Khosla, Aditya and Oliva, Aude and Torralba, Antonio},
  journal={IEEE Transactions on Pattern Analysis and Machine Intelligence},
  year={2017},
  publisher={IEEE}
}

@article{field1987relations,
  title={Relations between the statistics of natural images and the response properties of cortical cells},
  author={Field, David J},
  journal={Journal of the Optical Society of America A},
  volume={4},
  number={12},
  pages={2379--2394},
  year={1987},
  publisher={OSA}
}

@inproceedings{dhariwal2021diffusion,
  title={Diffusion models beat gans on image synthesis},
  author={Dhariwal, Prafulla and Nichol, Alexander},
  booktitle={NeurIPS},
  year={2021}
}

@inproceedings{ho2021classifier,
  title={Classifier-Free Diffusion Guidance},
  author={Ho, Jonathan and Salimans, Tim},
  booktitle={NeurIPS},
  year={2021}
}

\end{document}